\documentclass{article}
\usepackage{spconf,amsmath,graphicx, breqn, amssymb}
\usepackage{url}
\usepackage{units}

\DeclareMathOperator{\EX}{\mathbb{E}}
\newcommand\norm[1]{\left\lVert#1\right\rVert}

\newcommand\blfootnote[1]{%
  \begingroup
  \renewcommand\thefootnote{}\footnote{#1}%
  \addtocounter{footnote}{-1}%
  \endgroup
}

\title{Automatic Video Colorization using 3D Conditional Generative Adversarial Networks}
\name{Panagiotis Kouzouglidis $^{\star}$ \qquad Giorgos Sfikas$^{\star \dagger}$ \qquad Christophoros Nikou$^{\star}$ }

\address{$^{\star}$ Dpt. of Computer Science and Engineering,
University of Ioannina,
45110 Ioannina, Greece \\
    $^{\dagger}$ Information Technologies Institute,
    CERTH,
    57001 Thessaloniki, Greece}

\begin{document}
%
\maketitle
\begin{abstract}
  In this work, we present a method for automatic colorization of grayscale videos.
  The core of the method is a Generative Adversarial Network that is trained and tested on sequences of frames in a sliding window manner.
  Network convolutional and deconvolutional layers are three-dimensional, with frame height, width and time as the dimensions taken into account.
  Multiple chrominance estimates per frame are aggregated and combined with available luminance information to recreate a colored sequence.
  Colorization trials are run succesfully on a dataset of old black-and-white films.
  The usefulness of our method is also validated with numerical results, computed with a newly proposed metric that measures colorization consistency over a frame sequence.

   
\end{abstract}

\blfootnote{We gratefully acknowledge the support of NVIDIA Corporation with the donation of the Titan XP GPU used for this research.}

\begin{figure*}[!htb]
  \center{
    \begin{tabular}{c}
      \includegraphics[width=150mm, height=50mm]{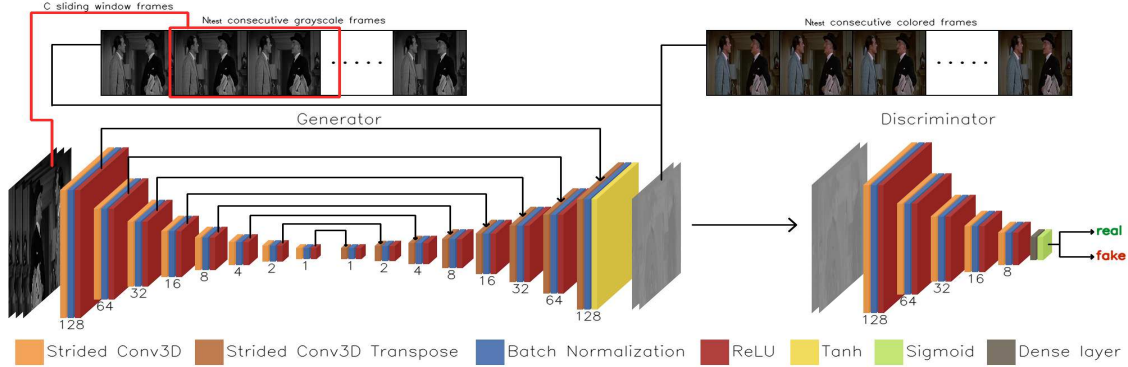} \\
    \end{tabular}
    \caption {\label{fig:proposed_architecture} \small{
    Architecture of the proposed model. 
    Sequences of luminance frames ('L' channels) are fed to the generator network, which creates chrominance channels for the corresponding frames ('a', 'b' channels).
    The discriminator network quantifies to what extend the generated chrominance sequence corresponds to a plausible colorization.
    As inputs and outputs are frame sequences, all convolutions are three-dimensional.
    During training, model weights are estimated through optimization of a GAN loss that effectively combines these constraints: 
    how plausible the generated colorization is per frame \emph{and as a sequence}, and how close the chrominance estimate is to the ground truth.
    In test time, the video is processed in a sliding window manner, producing $C$ colorization estimates (the size of the sliding window) for each frame,
    which are then aggregated to obtain a single estimate per frame.
    The output chrominance is finally combined with the luminance input in order to recreate a sequence of colorized frames. 
    } }
    \label{fig:model_architecture}
  }
\end{figure*}

\vspace{-0.8cm}
\section{Introduction}
\label{sec:intro}
In this paper, we address the problem of automatic colorization of monochrome digitized videos \cite{meyer2018deep,otani2014video,veeravasarapu2012fast,xia2016robust,yatziv2006fast}.
Perhaps the most straightforward practical application is to colorizing black-and-white footage from old films or documentaries.
Video compression is another possible application of note \cite{yatziv2006fast}.

Video colorization methods can be categorized according to the level of user interaction required.
A group of methods assume that a partially colored frame exists in the video, where color has been manually annotated in the form of color seeds \cite{yatziv2006fast,levin2004colorization,sheng2014video}. 
The method must then propagate color from these seeds to the rest of the frame, then to other frames in the video.
Other methods assume instead that a reference colored image exists that is similar in content and structure to the target monochrome video frames \cite{veeravasarapu2012fast,xia2016robust,ben2015approximate}.
These methods may or may not require user intervention; for example, in \cite{welsh2002transferring} the user can specify matching areas between the reference and the target frames.
In reference image-based methods, the problem of video colorization is hence converted to the problem of how to propagate color from the reference frame to other frames and/or from frame to frame.
Optical flow estimation has been used to guide frame-to-frame color propagation \cite{otani2014video,veeravasarapu2012fast}.
In \cite{sheng2014video}, Gabor feature flow is used as alternative to standard optical flow as a more robust guide to color propagation.
Naturally, methods of this vein work best for coloring short videos or frames coming from the same scene \cite{veeravasarapu2012fast}.

In the present work, we propose a learning-based method for video colorization.
As such, we assume that a collection of colored frames exist, that will be used to train the model.
In particular, the proposed method is based on an appropriately designed Generative Adversarial Network (GAN) \cite{goodfellow2014generative}.
GANs have gained a fair amount of traction in the last few years.
Despite their being harder to train even more than standard neural networks, requiring the employment of various heuristics and careful choosing of hyperparameters to attain convergence to a Nash equilibrium \cite{salimans2016improved,daskalakis2017traininggans},
they have proven to be excellent generative models.
The proposed model employs a conditional GAN (cGAN) architecture, popularized by the pix2pix model \cite{isola2016image}.
In the current work, convolutional and deconvolutional layers are 3D (height, width, time dimensions) to accomodate for the sequential nature of video data.

The main novel points of the current paper are as follows:
(a) we present a model for learning-based automatic video colorization that can take advantage of the sequential nature of video, while avoiding the use of frame-by-frame color propagation techniques that come with their own inherent limitations (typically they require existing colored key frames and/or are practically applicable within a single shot).
Other recent works use learning methods to color video via propagation \cite{meyer2018deep},
or via frame-by-frame image colorization, with each frame processed separately \cite{juliani2017};
(b) we elaborate on the issue of video colorization evaluation and propose a quantitative colorization metric specifically for video;
(c) we show that the proposed method creates colorization models that are transferable, in the sense that learning over a particular frame sequence produces a plausible output usable on a sequence of different content.

The rest of the paper is organized as follows.
In section \ref{sec:model}, we briefly discuss preliminaries on adversarial nets and present the architecture and processing pipeline of the proposed video colorization method.
In section \ref{sec:metrics}, we elaborate on existing numerical evaluation methods and propose a new metric to evaluate video colorization.
In section \ref{sec:experiments}, we show numerical and qualitative results of our method, tested on a collection of old films.
We close the paper with section \ref{sec:conclusion}, where we discuss conclusions and future work.



\begin{figure}[!htb]
  \center{
    \begin{tabular}{cccc}
    	\small{Grayscale} & \small{Ground Truth} & \small{Proposed} \\

      \includegraphics[width=22mm,height=20mm]{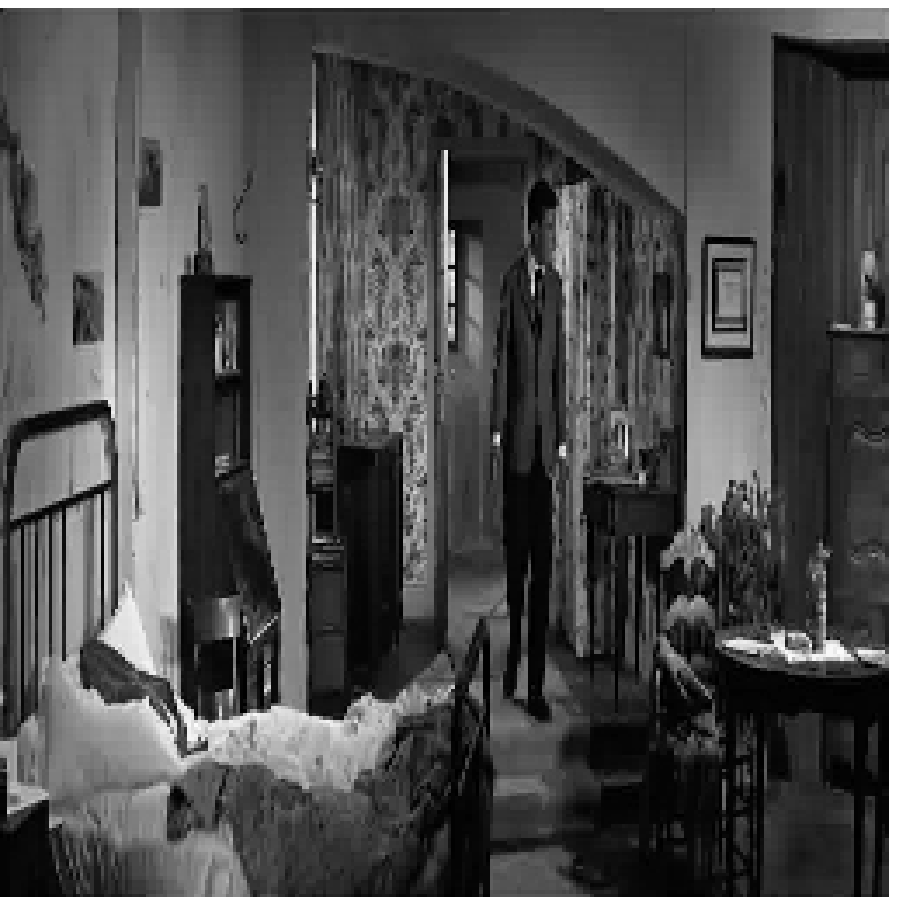} &
      \includegraphics[width=22mm,height=20mm]{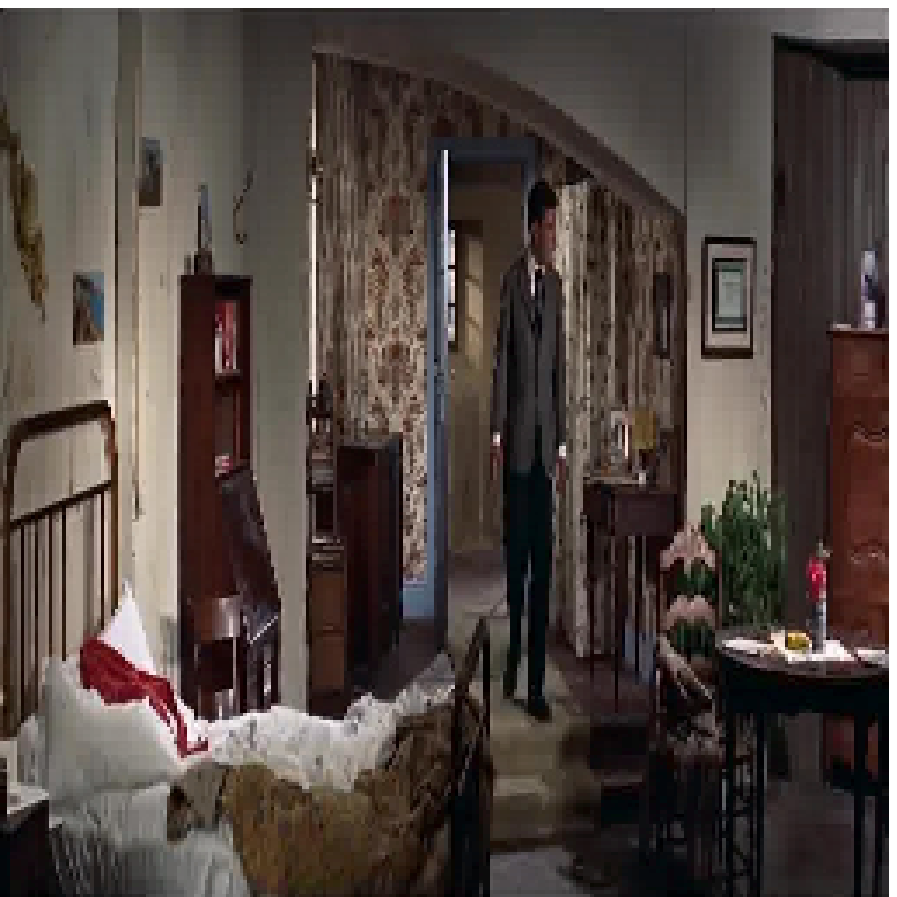} &
      \includegraphics[width=22mm,height=20mm]{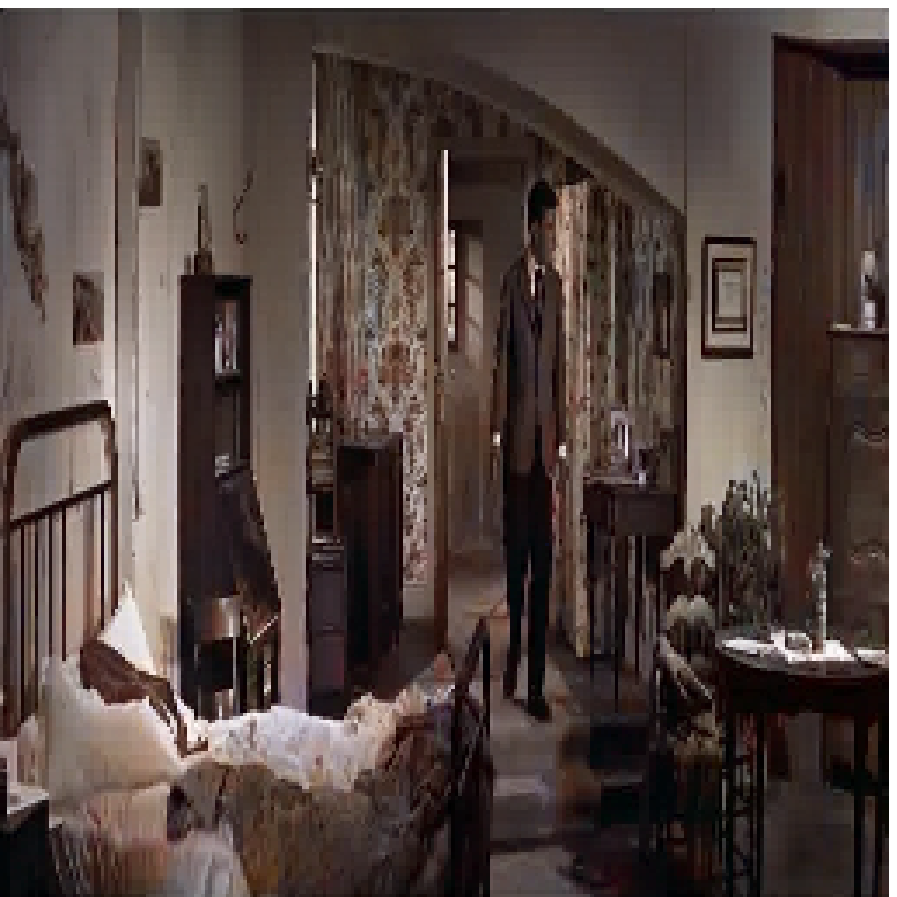} \\

      \includegraphics[width=22mm,height=20mm]{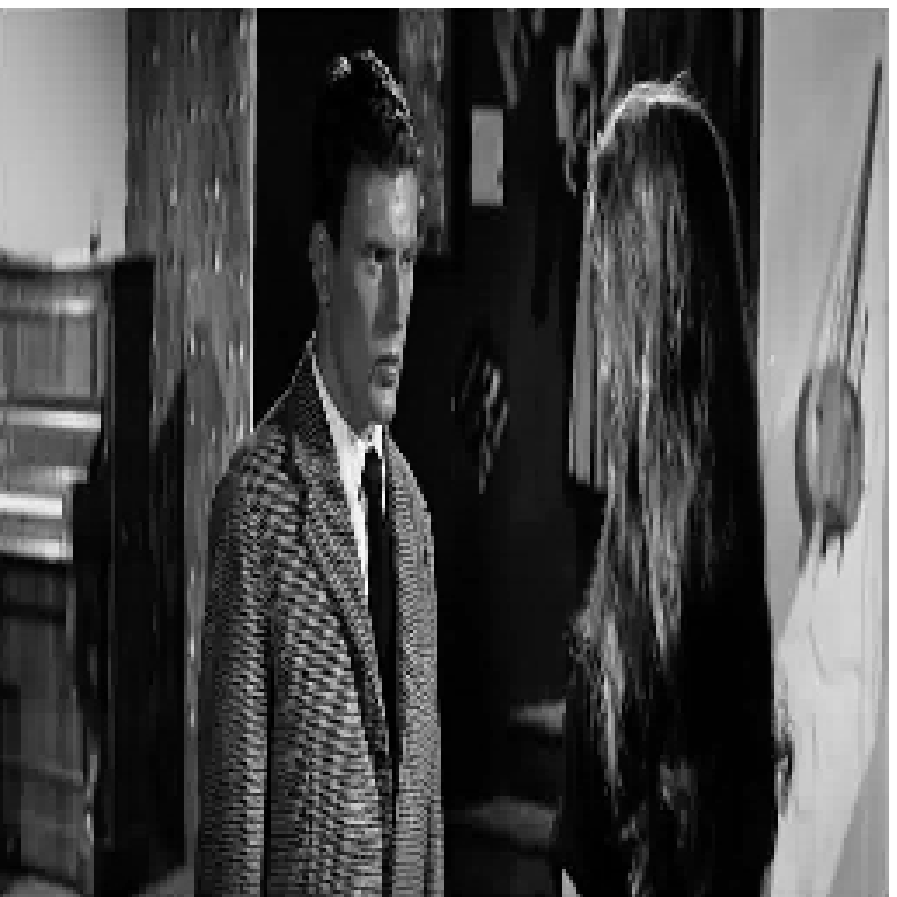} &
      \includegraphics[width=22mm,height=20mm]{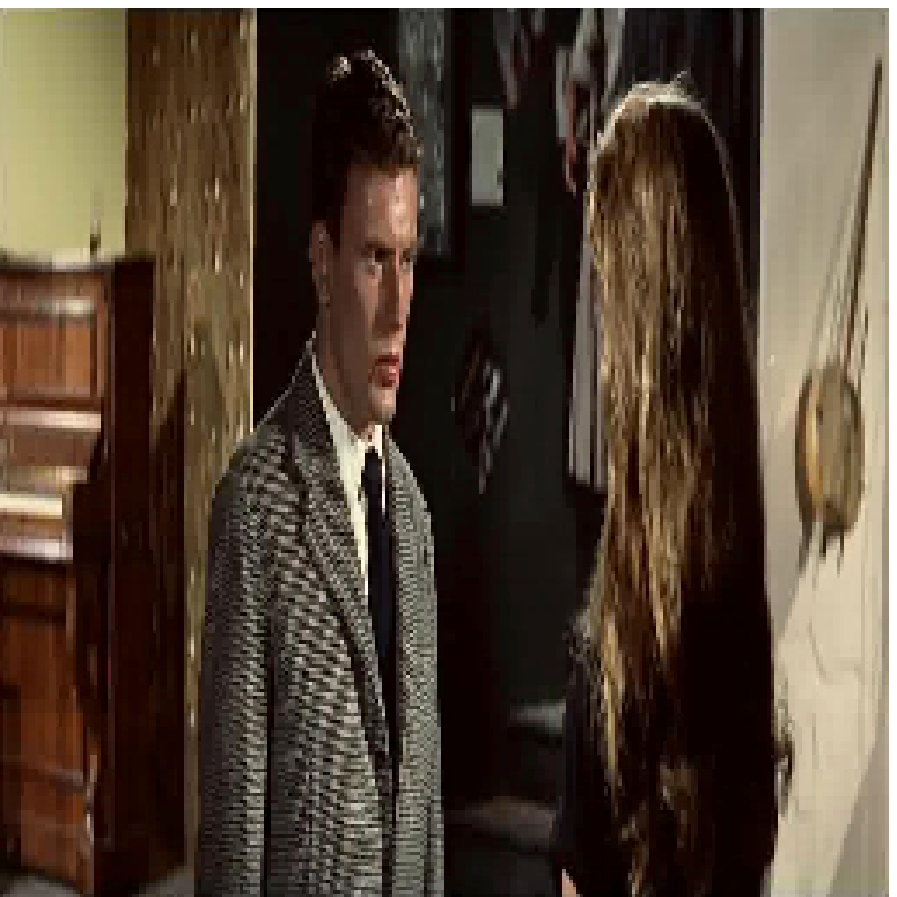} &
      \includegraphics[width=22mm,height=20mm]{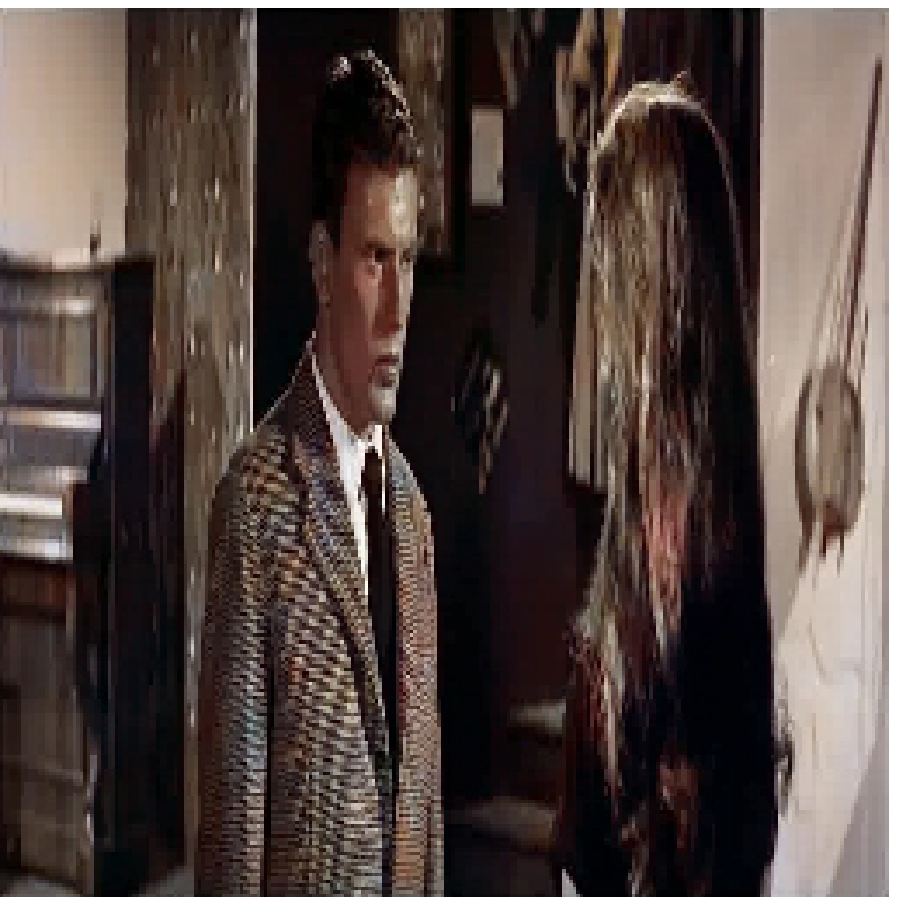} \\

      \includegraphics[width=22mm,height=20mm]{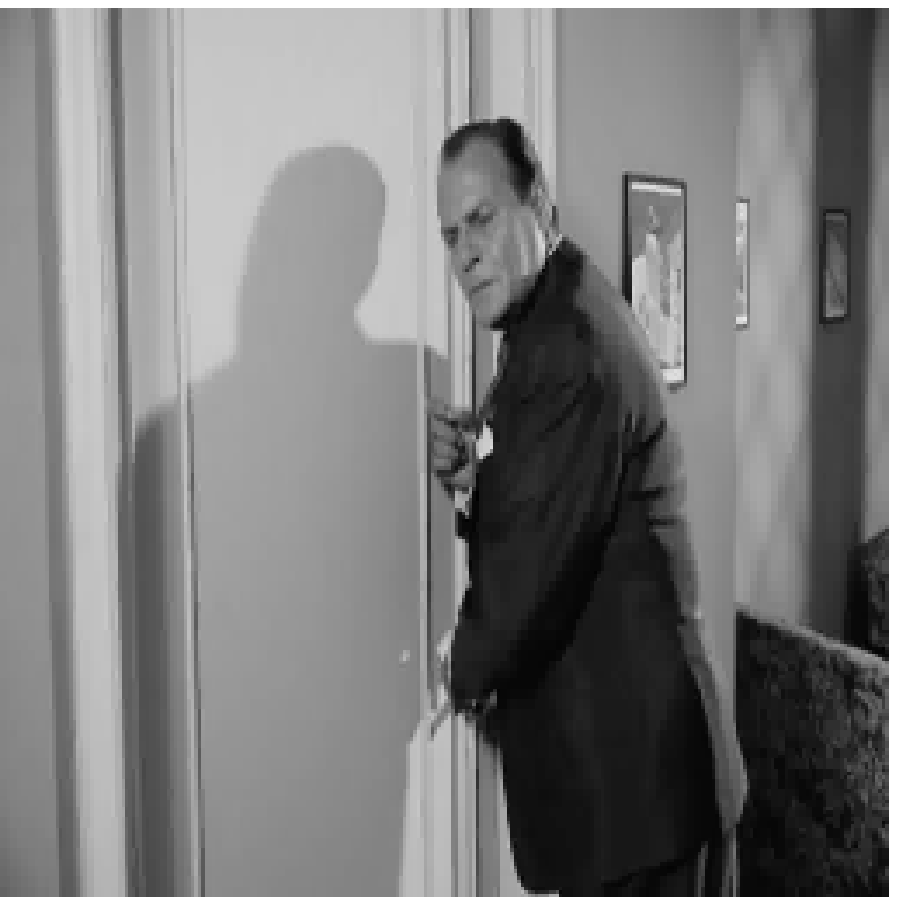} &
      \includegraphics[width=22mm,height=20mm]{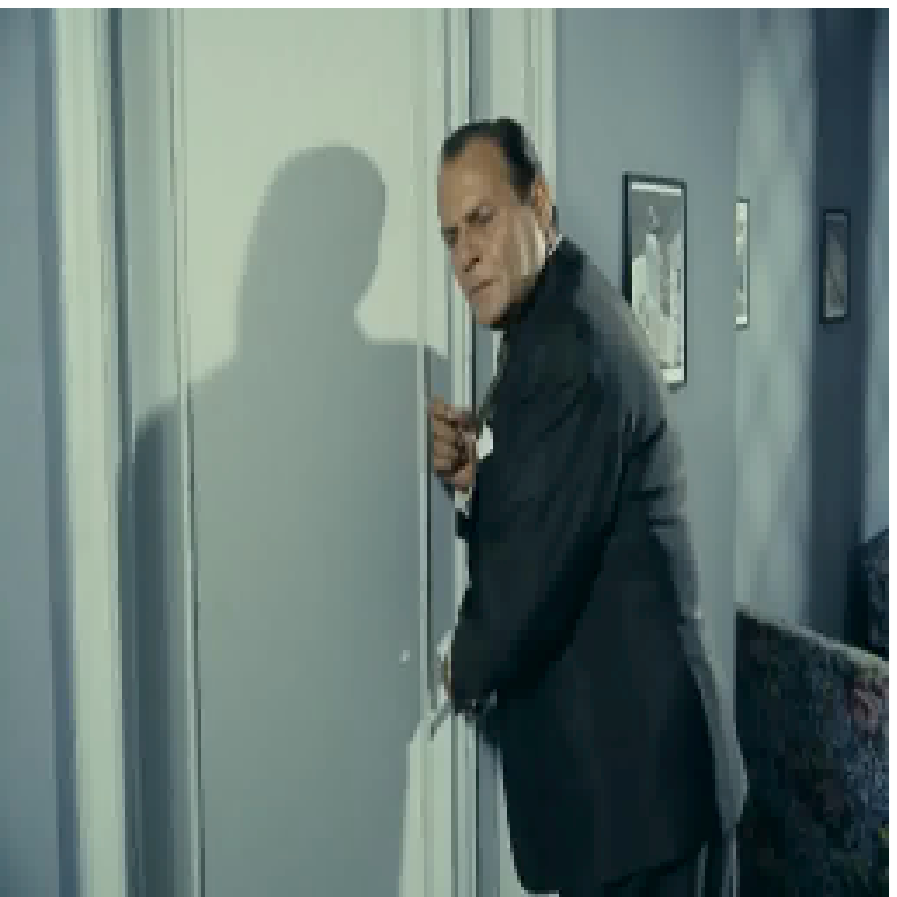} &
      \includegraphics[width=22mm,height=20mm]{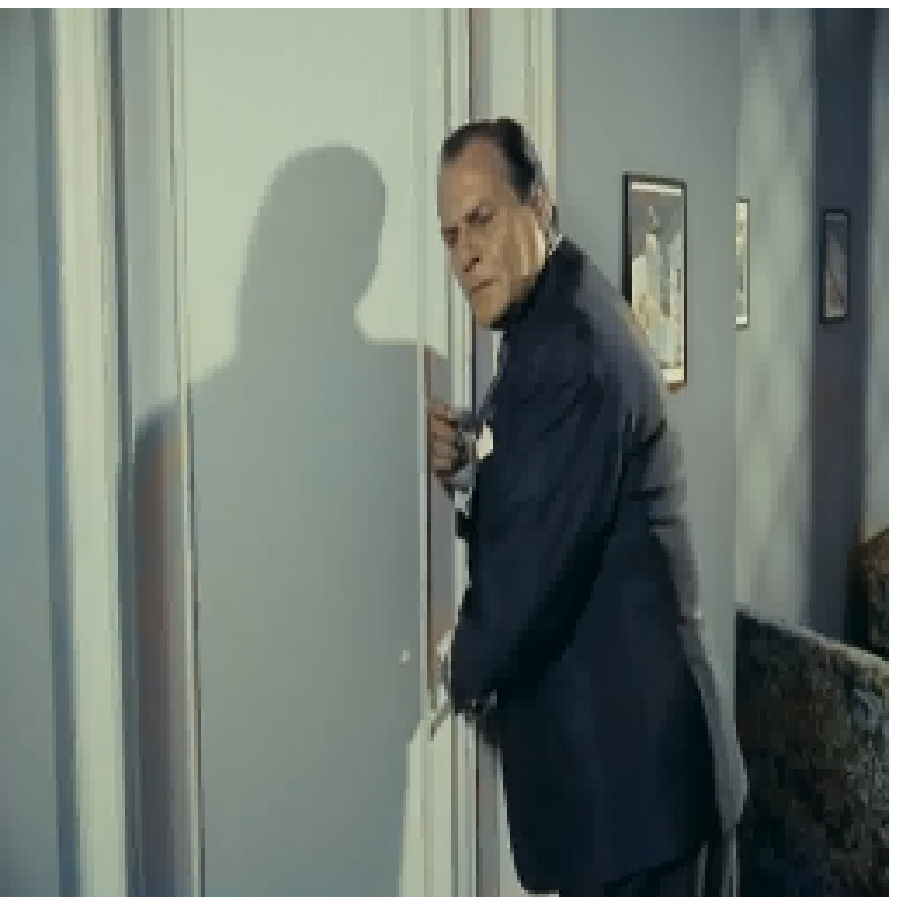} \\

      \includegraphics[width=22mm,height=20mm]{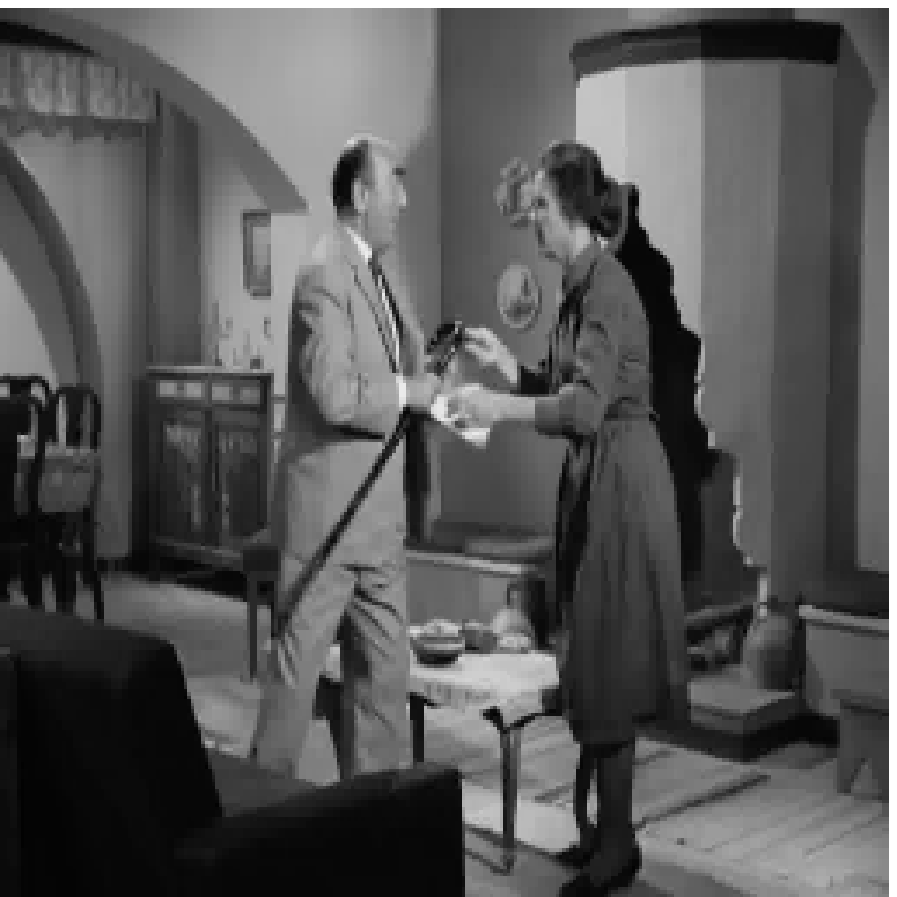} &
      \includegraphics[width=22mm,height=20mm]{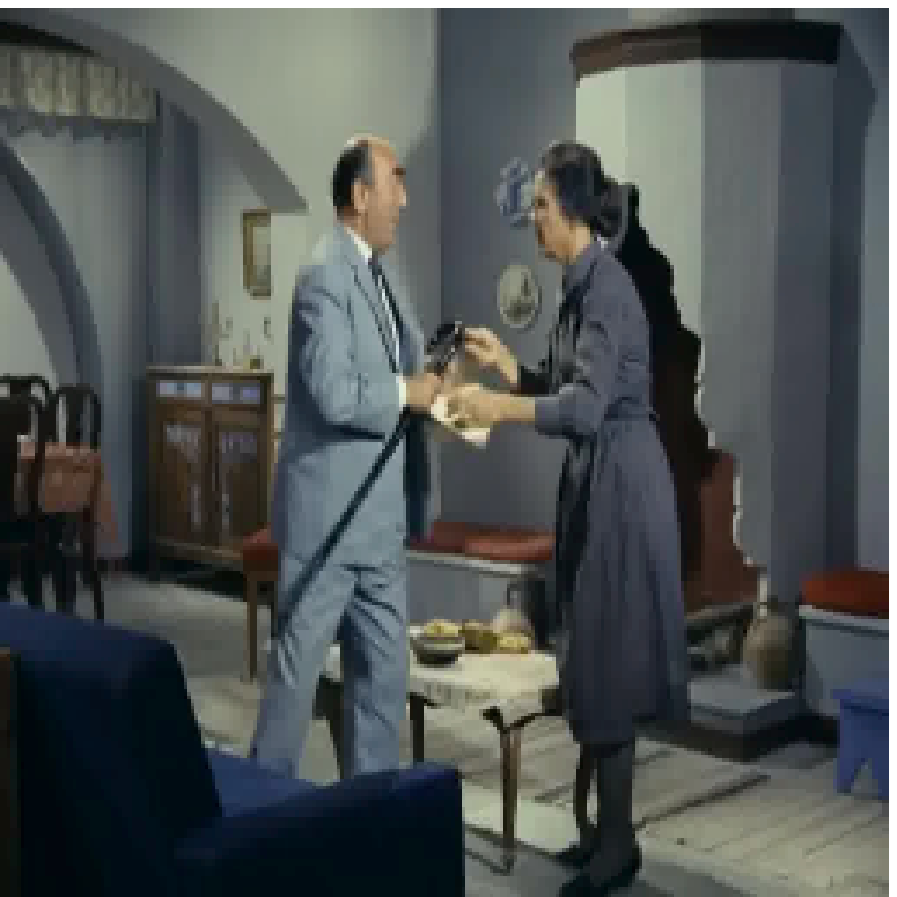} &
      \includegraphics[width=22mm,height=20mm]{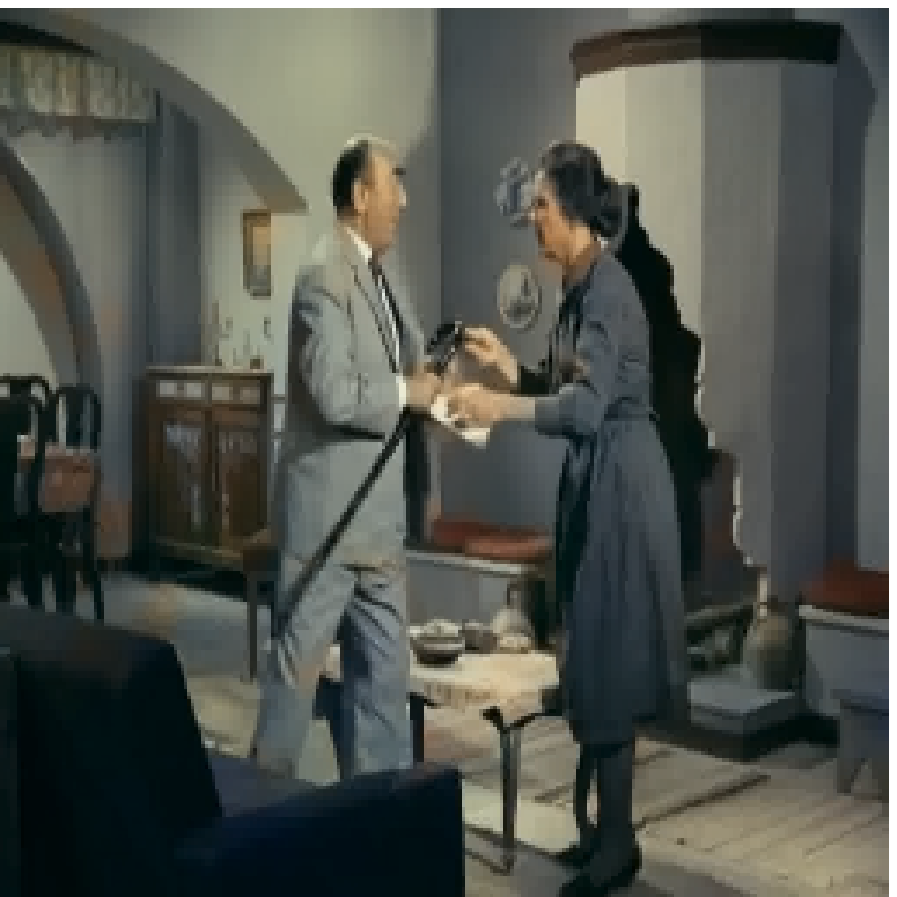} \\

    \end{tabular}
    \caption {\label{fig:qualitative_results1} \small{
    Colorization results using our method.
    Depicted frames are samples from films: ``Et Dieu..cr\'{e}a la femme'' and ``Tz\'{e}ni, Tz\'{e}ni'' (2 top and 2 bottom rows respectively).
    } }
  }
\end{figure}

\begin{figure}[!htb]
  \center{
    \begin{tabular}{cc}
      \small{2D cGAN} & \small{Proposed} \\ 
      \includegraphics[width=40mm,height=30mm]{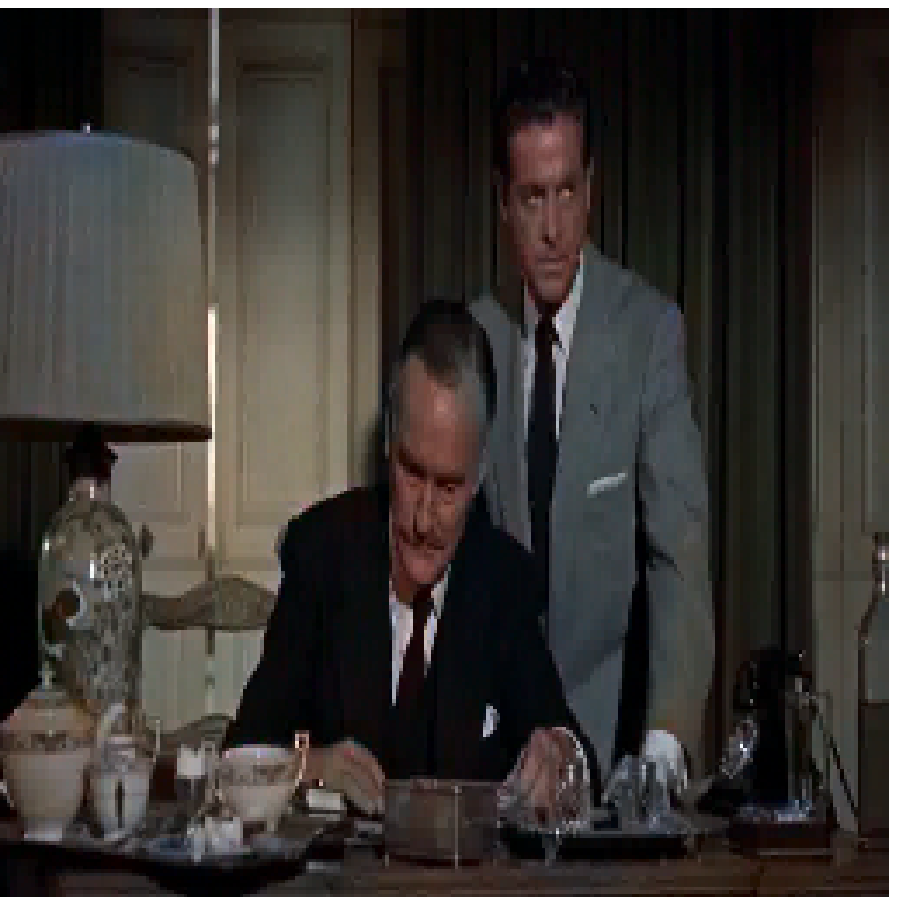} &
      \includegraphics[width=40mm,height=30mm]{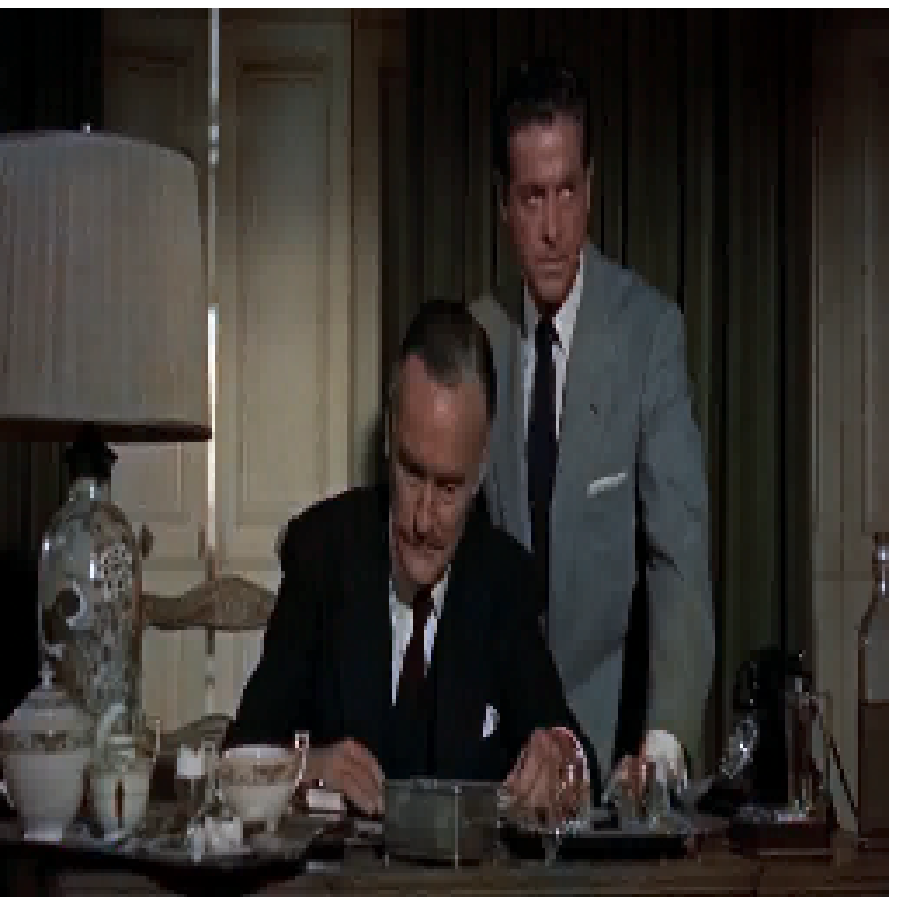} \\

      \includegraphics[width=40mm,height=30mm]{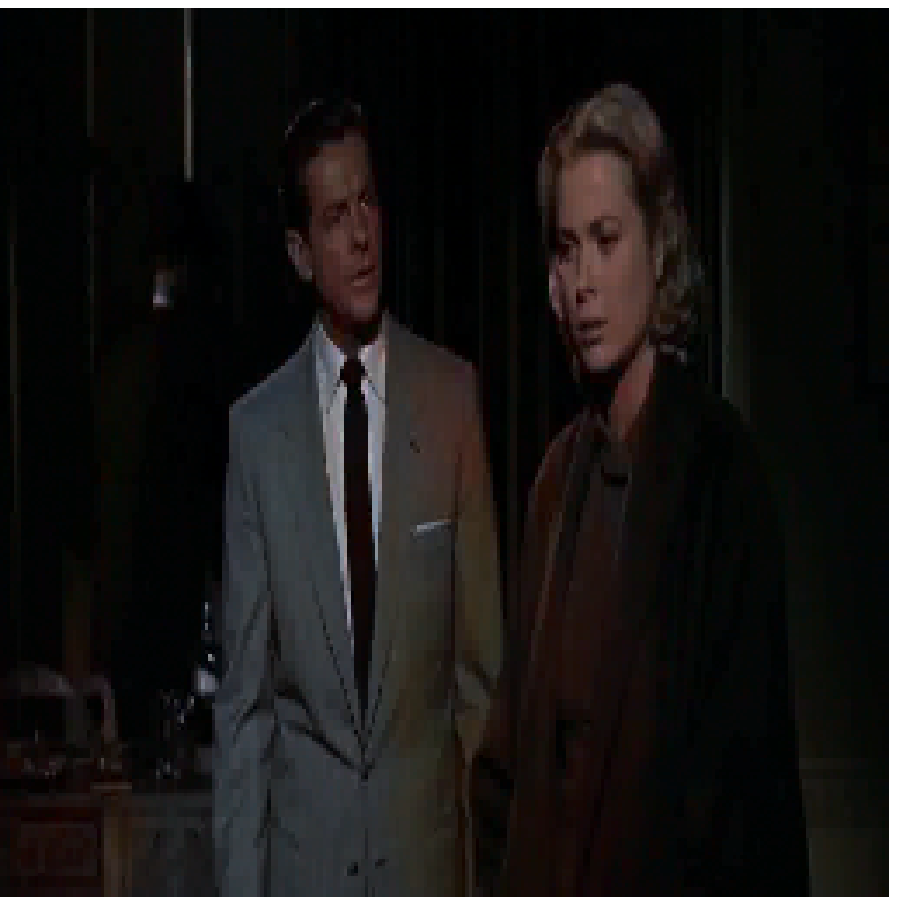} &
       \includegraphics[width=40mm,height=30mm]{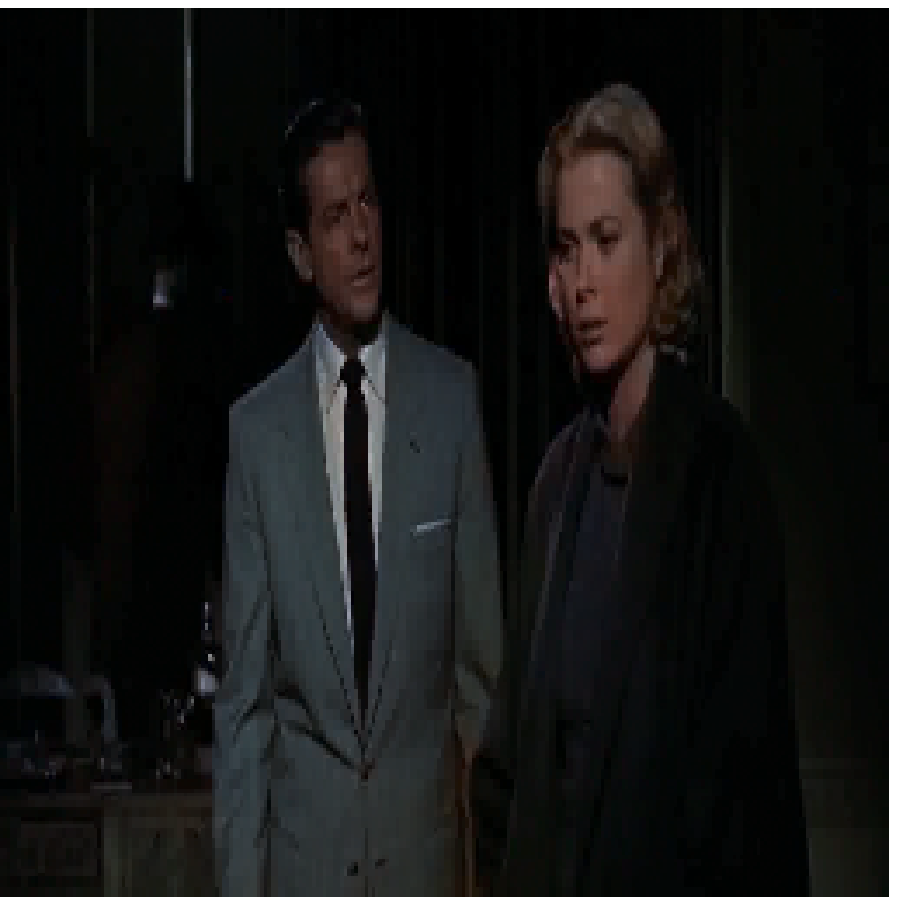} \\

    \end{tabular}
    \caption {\label{fig:qualitative_results_2Dvs3D} \small{
      Comparison of proposed model vs non-sequential 2D cGAN model.
    The proposed model produces better results than the non-sequential variant,
    as the former can take advantage of optical flow information, with its 3D convolution/deconvolution layers and estimate aggregation scheme.
    (Note for example how each method colorizes the hand of the standing actor on the top frame, or the color of the suit on the bottom frame).
    Depicted frames are samples from the film ``Dial M for Murder''.
    } }
  }
\end{figure}


\vspace{-0.3cm}
\section{Proposed method}
\label{sec:model}

The proposed method assumes the existence of a training set consisting of a sequence of $N_{train}$ colored frames,
and a test set consisting of a sequence of $N_{test}$ monochrome frames that are to be colorized.
During the training phase, a cGAN model is used to learn how to color batches of $C$ ordered frames.
Hyperparameter $C$ is fixed beforehand with $C << min\{ N_{train},N_{test}\}$. 
During the testing phase, the model is run on the input monochrome video in a sliding window manner.
Windows are overlapping and move by a single frame at a time, thereby producing a set of $C$ colorization estimates \emph{for each} monochrome frame, hence $C$ video colorization proposals.
These $C$ estimates are then combined to produce a single colorized output.
In what follows we discuss the details of this process.

A GAN is a generative, neural network-based model that consists of two components, the generator network and the discriminator network.
The cGAN architecture \cite{isola2016image} that is employed as part of the proposed method, is a supervised variant of the original unsupervised GAN \cite{goodfellow2014generative}.
A cGAN learns a mapping from observed input $x$ to target output $y$
\footnote{Other variants of a cGAN are possible; for example, a noise variable $z$ could be added to produce a non-deterministic output \cite{isola2016image}.
We employ a deterministic cGAN variant in this work.}.

Formally, the objective to be optimized is:

\[
\arg\min_{G} \max_{D}[\EX_{y}[logD(y)]+
\]
\vspace{-0.5cm}
\begin{dmath}
  +\EX_{x}[log(1 - D(G(x)))]] + \lambda\EX_{x, y}[\norm{y - G(x)}_1],
\end{dmath}
with hyperparameter $\lambda$ controlling the trade-off between the GAN (discriminator) loss and the $L_1$ loss.
$D(\cdot)$ and $G(\cdot)$ correspond to the discriminator and generator respectively.
The GAN loss quantifies how plausible the colorization output, while the $L_1$ loss forces the colorization to be close to the ground truth.
We use representations in the CIE Lab color space (following e.g.\cite{zhang2016colorful}).
For a monochrome frame sequence, only luminance is known beforehand.
Input is a sequence of luminance channels (channel $L$) of \textit{C} consecutive frames $x \in$ ${\rm I\!R}^{H \times W \times C \times 1}$, 
and the objective is to learn a mapping from luminance to chrominance (channels $a$,$b$) $y \in$ ${\rm I\!R}^{H\times W \times C \times 2}$ where \textit{H}, \textit{W} are frame dimensions.

The generator network is comprised of a series of convolutional and deconvolutional layers. 
Skip connections are added in the manner introduced by UNet \cite{ronneberger2015u}.
As inputs and outputs are sequences of fixed-size frames, all convolutions and deconvolutions are three-dimensional (frame height, width and time dimensions).
The encoder and decoder stacks comprise $8$ strided convolutional/deconvolutional layers each (stride=$2$), followed iteratively by batch normalization (BN) and rectified linear unit (ReLU) activation layers.
Following \cite{chintala2016}, outputs are forced to lie in the $(-1,1)$ range with a tanh activation layer at the end of the generator network, and only later renormalized to valid $a,b$ chrominance values.
The discriminator network is a 3D convolutional network comprising $5$ convolutional layers iteratively followed by BN and ReLU layers.
The discriminator is topped by a fully connected (``dense'') layer and a sigmoid activation unit in order to map the image to a real/fake probability figure.

At test time, we use the generator in a sliding window fashion over the footage to be colorized. 
Hence, each frame is given as input to the generator at a total of $C$ times, since $C$ is the size of the sliding window.
The produced $C$ chrominance estimates $\chi_1, \chi_2, \cdots, \chi_C$
\footnote{$\chi_i$ denotes the $i^{th}$ colorization estimate for a frame. $y$ denotes a colorization estimate for a sequence of $C$ frames.}
then need to be used to produce a single estimate $\hat{\chi}$.
We can write $\hat{\chi}$ as a maximum-a-posteriori (MAP) estimate as:
\begin{equation}
  \label{eq:aggregation}
\hat{\chi} = \arg\max_{\chi} p(\chi|\chi_1,\cdots,\chi_C) 
\end{equation}
where a prior distribution $p(\chi)$ can be assumed over possible $a,b$ values in order to favor a particular chrominance setup.
If identical distributions centered around each $\chi_i$ and an uninformative prior is used, the above formula simplifies as an average over all chrominance values per frame pixel:
$\hat{\chi} = \nicefrac{1}{C}\sum_{i=1}^C\chi_i$.
Finally, the chrominance estimate is recombined with input luminance to recreate colored RGB frames for the input video.
The architecture of the proposed model is summarized in figure ~\ref{fig:model_architecture}.

\vspace{-0.3cm}
\section{Metrics for numerical evaluation of video colorization}
\label{sec:metrics}




In this section we describe the metrics we use for numerical evaluation of video colorization.
We use two metrics that measure per-frame colorization quality, also usable in single-image colorization.
Furthermore, we propose a new metric suitable for video colorization in particular.

\emph{Peak Signal-to-Noise Ratio (PSNR):}
PSNR is calculated per each test frame in the RGB colorspace, and their mean is reported as a benchmark over the whole video.

\emph{Raw Accuracy (RA):}
Raw Accuracy, used in \cite{zhang2016colorful} to evaluate image colorization, is defined in terms of accuracy of predicted colors over a varying threshold.
Colors are classified as correctly predicted if their Euclidean distance in the $ab$ space is lower than a threshold.
Accuracy is computed over color values for every pixel position and frame.
Integrating over the curve that is produced by taking into account varying threshold yields the RA metric.
We integrated from $0$ to $150$ distance units as in \cite{zhang2016colorful}.

\emph{Color Consistency (CC):} 
The aforementioned metrics measure strictly the quality of colorization of each frame separately.
We propose and use a metric to measure both per-frame quality and also the consistency of the choice of colors between consecutive frames.
Such a metric can, for example, penalize erratic differences in colorization from frame to frame,
that would otherwise be ``invisible'' to the other metrics, borrowed from single image restoration/colorization.
We define color consistency over sets of two consecutive colorization predictions $\hat{\chi}^{(t)}, \hat{\chi}^{(t+1)}$ and corresponding ground truth values $\chi^{(t)},\chi^{(t+1)}$ as
\begin{equation}
  CC^{(t,t+1)} = \nicefrac{1}{HW} \sum_{i=1}^H \sum_{j=1}^W \nicefrac{1}{2}[A^{(t)}_{ij} + A^{(t+1)}_{ij}] A^{(t \times t+1)}_{ij}
\end{equation}
where affinity matrices $A^{(t)}$ and $A^{(t \times t+1)}$ are defined as
\[
  A_{ij}^{(t)} = \phi(||\chi_{ij}^{(t)} - \hat{\chi}_{ij}^{(t)} ||),
\]
\[
  A_{ij}^{(t \times t+1)} = \phi (\left\lVert ||\chi_{ij}^{(t)} - \chi_{ij}^{(t+1)} || - ||\hat{\chi}_{ij}^{(t)} - \hat{\chi}_{ij}^{(t+1)} || \right\rVert),
\]
with function $\phi(\cdot)$ a positive, strictly decreasing function that is used to convert distances to similarities.
We use $\phi_{ij}(X) = \lfloor 60 X_{ij}/(\max(X)+\epsilon) + 1 \rfloor^{-1}$.
Total CC over a video sequence is calculated as the average CC over all consecutive frames.
Higher values correspond to better results.

\vspace{-.3cm}
\section{Experiments}
\label{sec:experiments}

We have tested our method over a collection of old films:
$(a)$ ``Dial M for Murder'' (USA, 1954; 63,243 frames) \cite{dialm}
$(b)$ ``Et Dieu..cr\'{e}a la femme'' (France, 1956; 54,922 frames) \cite{etdieu}
$(c)$ ``Tz\'{e}ni, Tz\'{e}ni'' (Greece, 1965; 58,932 frames) \cite{tzeni}
$(d)$ ``A streetcar named desire'' (USA, 1951; 18,002 frames) \cite{astreetcar}
$(e)$ ``Twelve angry men'' (USA, 1957; 12,000 frames) \cite{twelveangrymen}.
Frames were sampled off these films at $10$ fps.
Films $(a),(b),(c)$ are colored, while $(d)$ and $(e)$ are originally black-and-white.
Consequently, only the colored films could be used for training, while the black-and-white ones could be used only for testing with a colorizer trained on another film.

We have first experimented with training and testing on different parts of the same (colored) film.
For training/testing we have used the first 75\%/last 25\% from each of the colored films.
The proposed 3D cGAN model was used, with model parameters set to $C=3$ (sliding window size), $\lambda=100$ (GAN-$L_{1}$ loss tradeoff),
and compared against a 2D cGAN model that learned to colorize each frame separately.
We have also use data augmentation on our training set, with random horizontal flips ($50\%$ chance to use a flipped input during training) and gaussian additive noise ($\sim\mathcal{N}(0, 1.2e-3)$).
For estimate aggregation (eq.~\ref{eq:aggregation}) we present results with an uninformative prior 
(preliminary tests with priors learned over data statistics did not give any definite improvement).
We also compare with a greyscale baseline, i.e. the case where the ``colorized'' video estimate uses only luminance information.
Numerical results can be examined in table \ref{table:numerical_basic}.
Qualititative results can be examined in figures \ref{fig:qualitative_results1} and \ref{fig:qualitative_results_2Dvs3D}.
While in general both models fare satisfactorily, the proposed model can avoid erroneous colorizations in several cases (cf. fig. \ref{fig:qualitative_results_2Dvs3D}).
This point is validated by our numerical results, where we calculate the metrics presented in section \ref{sec:metrics}.
While w.r.t. to PSNR and RA the proposed model still is better, it could be argued that the difference in the result is statistically insignificant.
This is not the case with the proposed CC metric however, where the performance of the proposed model is markedly better.
These results validate our expectation, as the 3D structure of the proposed model can take into account the sequential structure of the video, in contrast to its 2D counterpart.

\begin{figure}[!htb]
  \center{
   \begin{tabular}{cccc}
      \includegraphics[width=18mm]{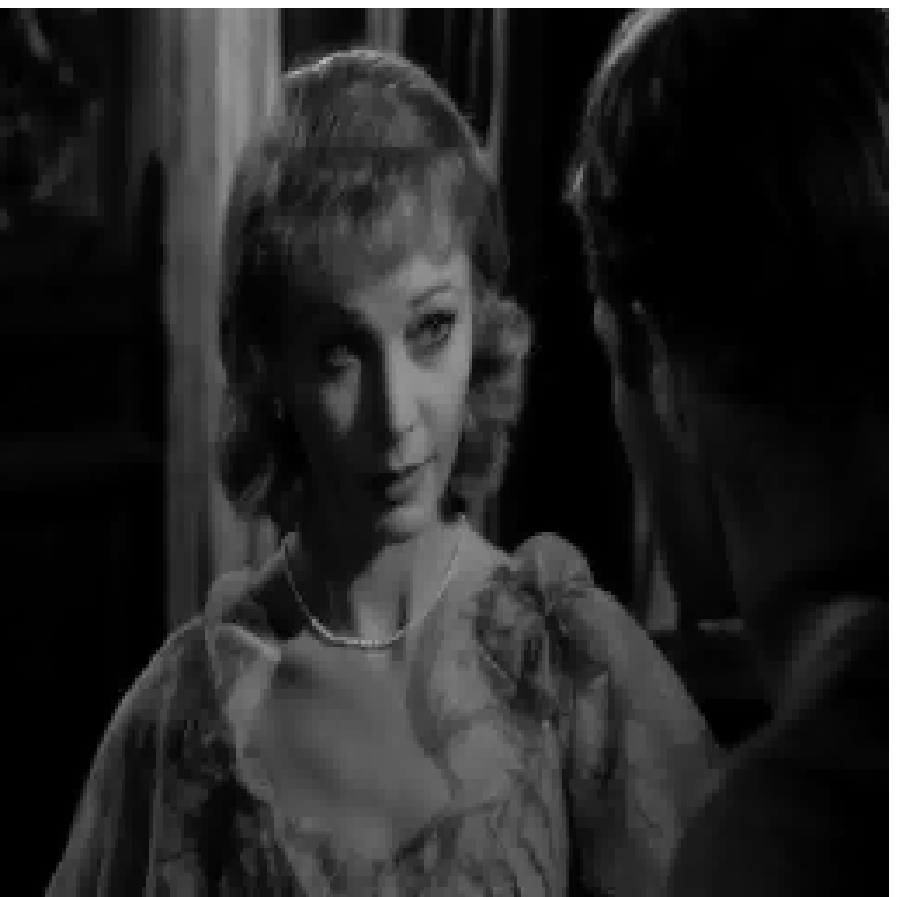} &
      \includegraphics[width=18mm]{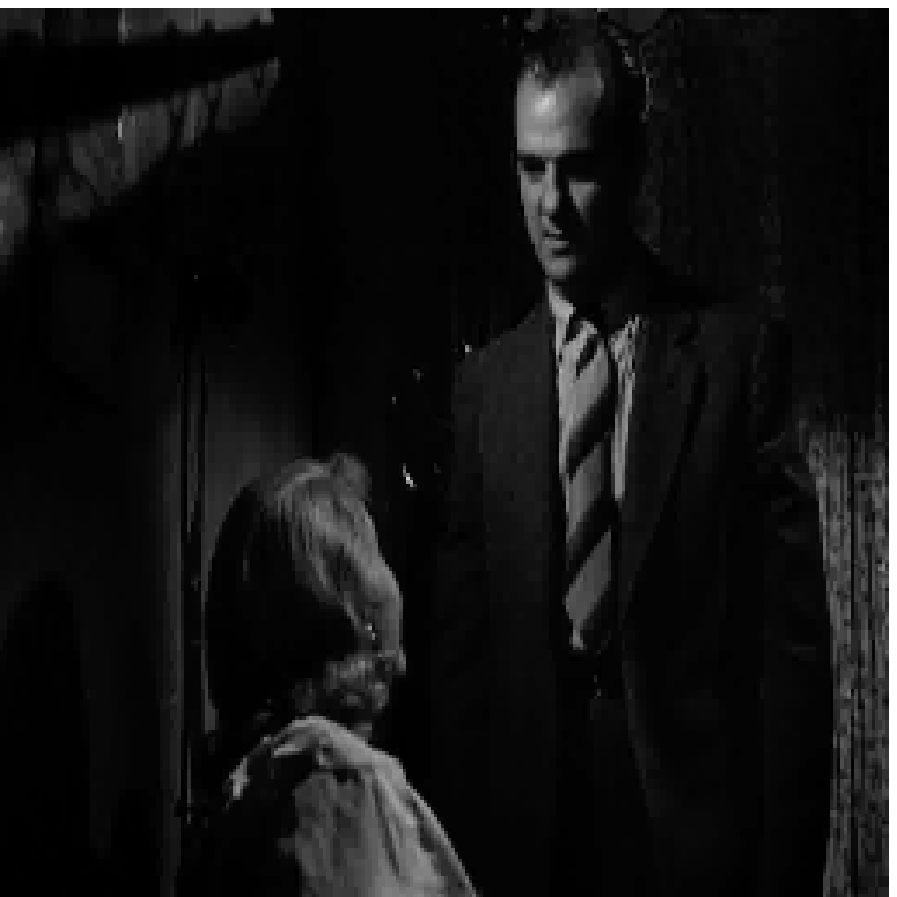} &
      \includegraphics[width=18mm]{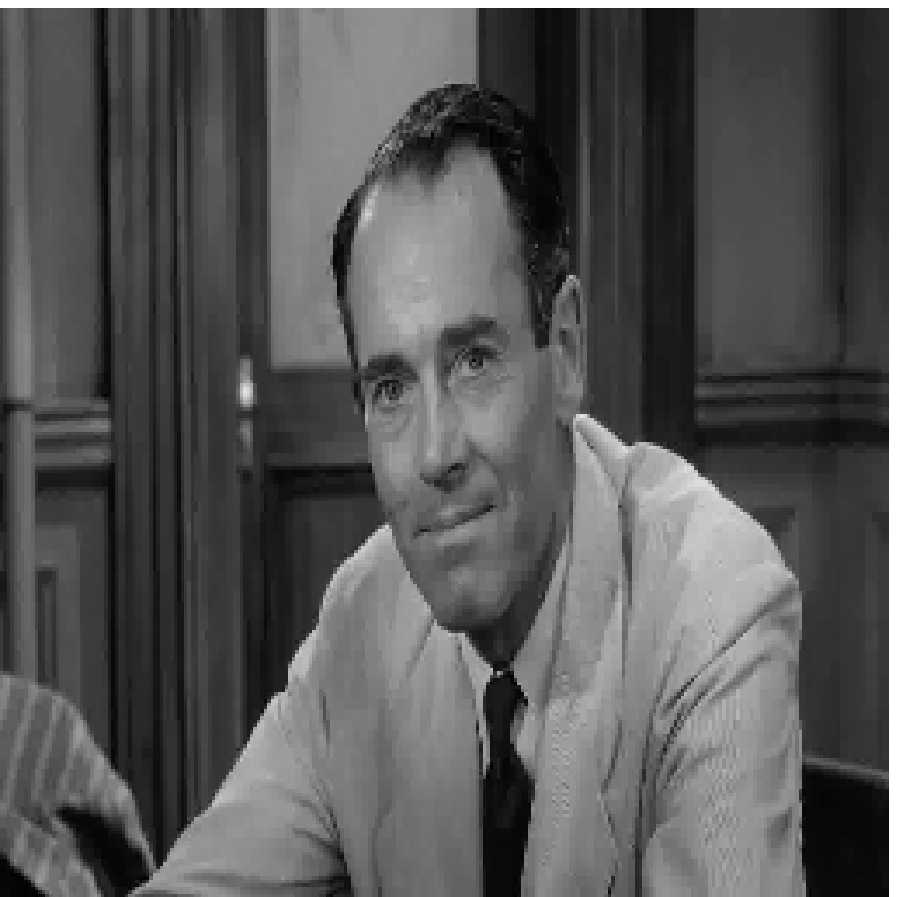} &
      \includegraphics[width=18mm]{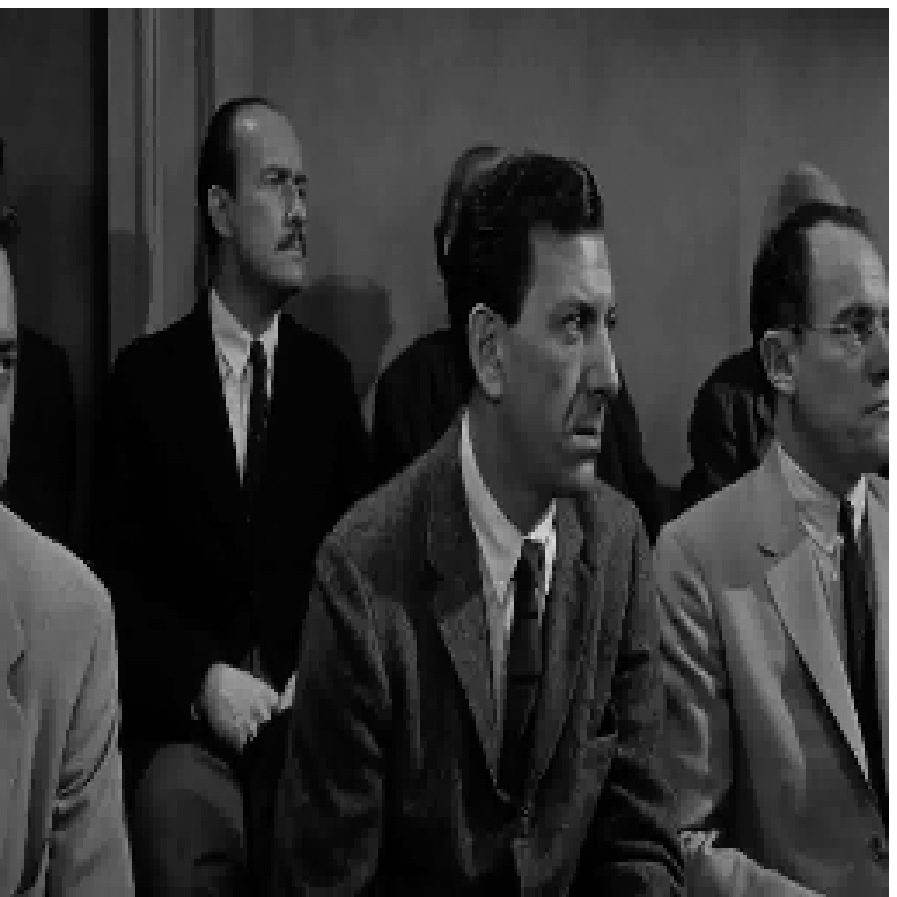} \\

      \includegraphics[width=18mm]{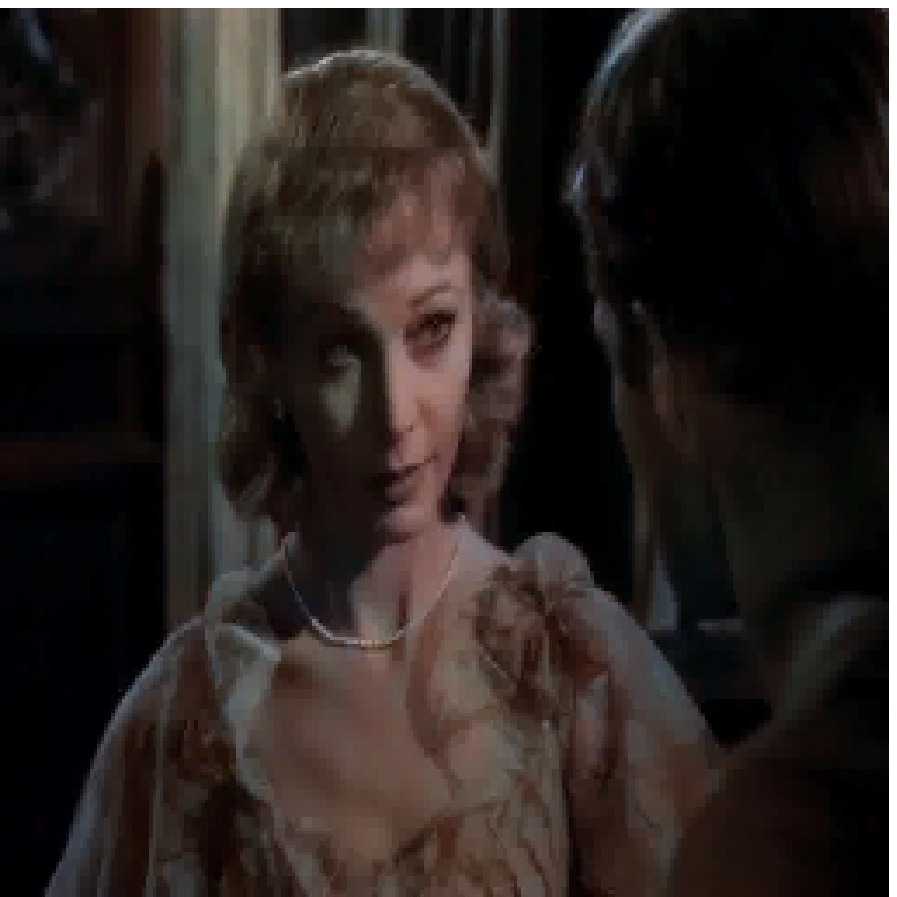} &
      \includegraphics[width=18mm]{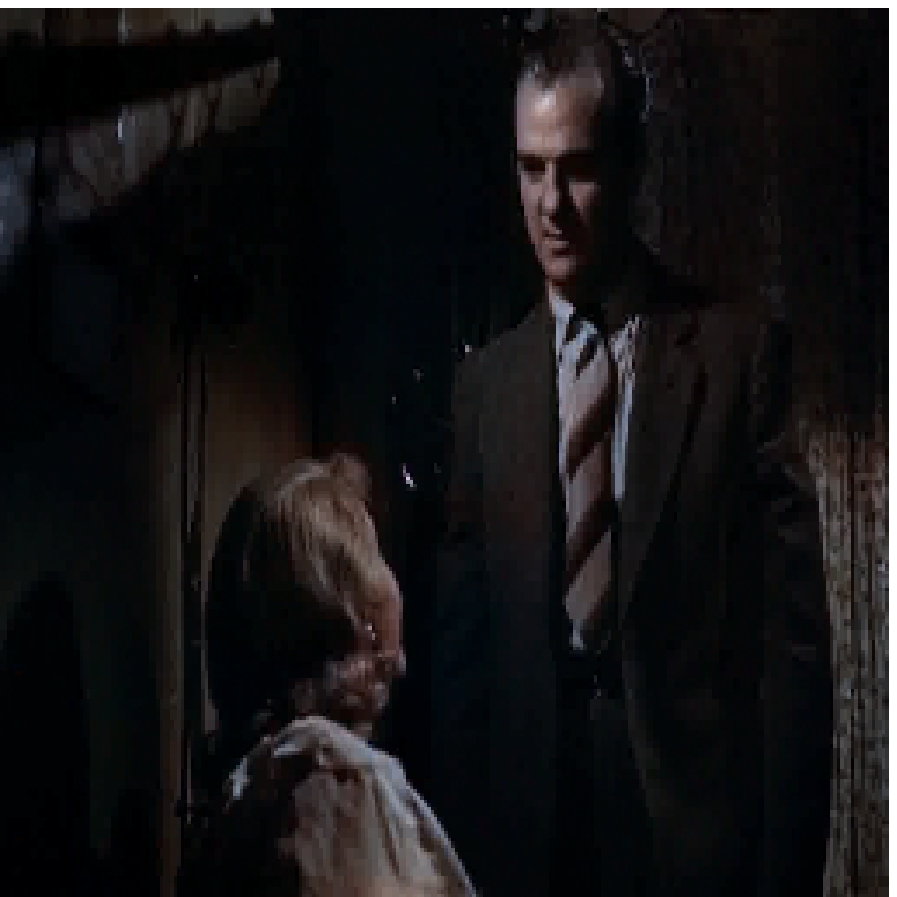} & 
      \includegraphics[width=18mm]{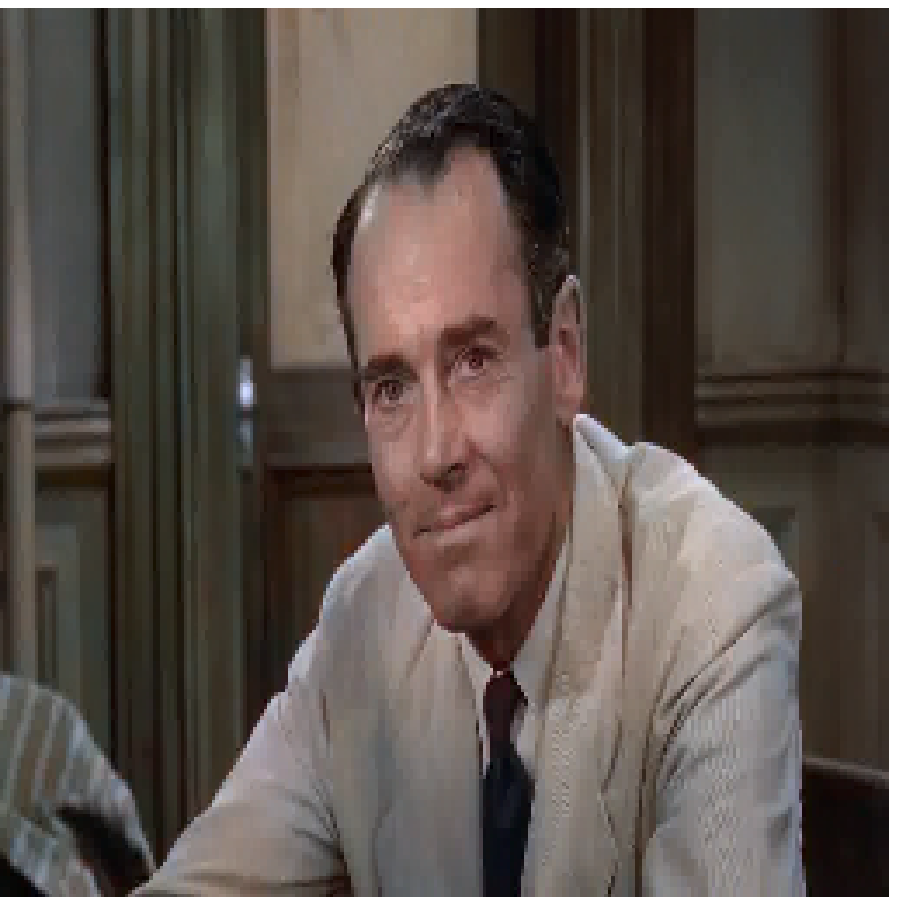} &
      \includegraphics[width=18mm]{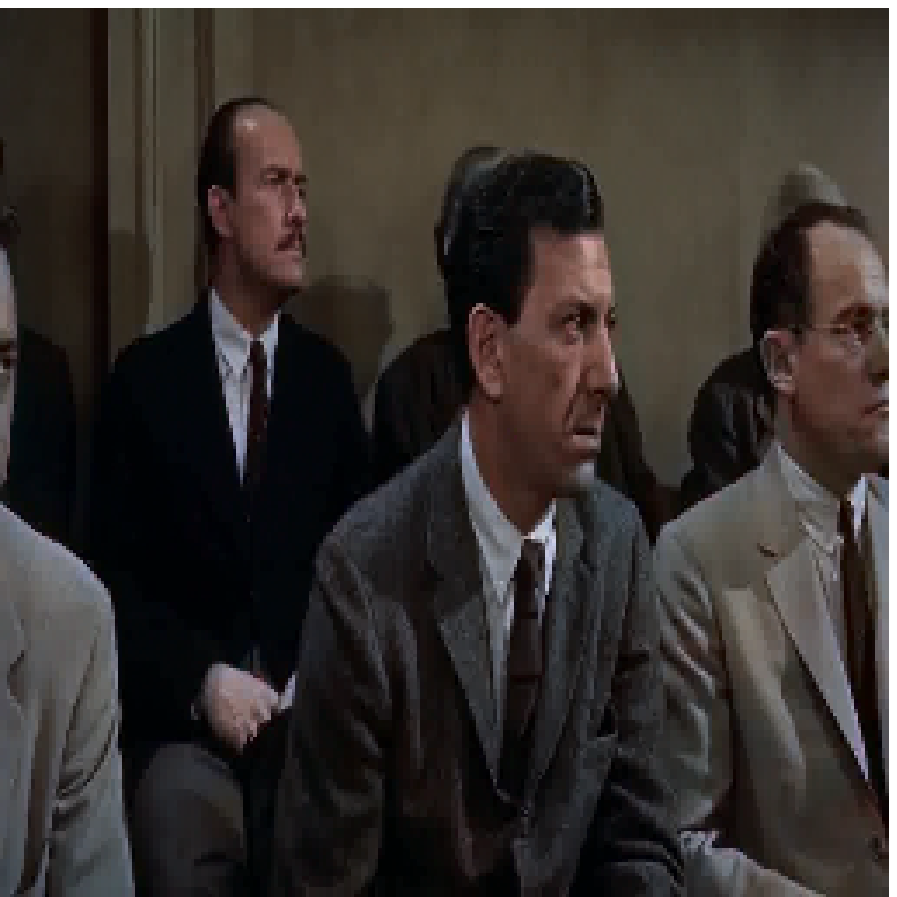} \\
    \end{tabular}
    \caption {\label{fig:qualitative_results2} \small{
   Colorization results where color ground-truth is unavailable.
   Depicted are samples from ``A streetcar named desire'' and ``Twelve angry men'' (2 leftmost, 2 rightmost columns respectively), colorized with the proposed model trained on ``Dial M for Murder''.
   } } 
  }
\end{figure}

We have also run tests for training and testing on different films.
The case that is perhaps closest to a practical application of the current model is using trained models on one of the colored films to color black-and-white footage, i.e. in our case films $(d)$ and $(e)$.
Results for this case can be examined at fig.~\ref{fig:qualitative_results2} (training performed on film $(a)$).
Video colorization demos are available online
\footnote{\url{http://www.cs.uoi.gr/~sfikas/video_colorization} }
.


\begin{table}[!htb]
  \center{
  \begin{tabular}{|c|c|c|c|}
	\hline
	 {} & PSNR & RA & CC \\ \hline \hline
	\multicolumn{4}{|c|}{(a) ``Dial M for Murder''} \\
	\hline
	\hline
	Grayscale & 32.69 & 96.55 & 73.09 \\
	\hline
	2D cGAN & 34.97 & 96.67 & 82.07 \\
	\hline
	Proposed& \textbf{35.66} & \textbf{96.73} & \textbf{85.59} \\
	\hline
	\hline
	\multicolumn{4}{|c|}{(b) ``Et Dieu..cr\'{e}a la femme''} \\
	\hline
	\hline
	Grayscale & 30.23 & 94.07 & 47.82 \\
	\hline
	2D cGAN  & 32.08 & 95.17 & 56.67 \\
	\hline
	Proposed & \textbf{32.32} & \textbf{95.31} & \textbf{58.80} \\
	\hline
	\hline
	\multicolumn{4}{|c|}{(c) ``Tz\'{e}ni, Tz\'{e}ni''} \\
	\hline
    \hline
	Grayscale & 29.83 & 92.85 & 39.17 \\
	\hline
	2D cGAN & 31.44 & 93.87 & 50.81 \\
	\hline
	Proposed & \textbf{31.77} & \textbf{94.14} & \textbf{55.16} \\
	\hline

    \end{tabular}
    \caption {\label{table:numerical_basic} \small{Numerical results for colorization evaluation.
    Training and testing is performed on different clips of the same film.
    PSNR is measured in dB; RA and CC values are percentages.
    Higher values are better.
    The proposed model performs best, in all cases.}}
  }
\end{table}

\vspace{-0.6cm}
\section{Conclusion and Future work}
\label{sec:conclusion}




We have presented a method for automatic video colorization, based on a novel cGAN-based model with 3D convolutional and deconvolutional layers and an estimate aggregation scheme.
The usefulness of our model has been validated with tests on colorizing old black-and-white film footage.
Model performance has also been evaluated with single-image based metrics as well as a newly proposed metric that measures sequential color consistency.
As future work, we envisage exploring the uses of the color prior in our aggregation scheme.


\clearpage
\bibliographystyle{IEEEbib}
\bibliography{refs}

\begin{thebibliography}{10}

\bibitem{meyer2018deep}
Simone Meyer, Victor Cornill{\`e}re, Abdelaziz Djelouah, Christopher Schroers,
  and Markus Gross,
\newblock ``Deep video color propagation,''
\newblock {\em arXiv preprint arXiv:1808.03232}, 2018.

\bibitem{otani2014video}
Mayu Otani and Hirohisa Hioki,
\newblock ``Video colorization based on optical flow and edge-oriented color
  propagation,''
\newblock in {\em Computational Imaging XII}. International Society for Optics
  and Photonics, 2014, vol. 9020, p. 902002.

\bibitem{veeravasarapu2012fast}
VS~Rao Veeravasarapu and Jayanthi Sivaswamy,
\newblock ``Fast and fully automated video colorization,''
\newblock in {\em Signal Processing and Communications (SPCOM), 2012
  International Conference on}. IEEE, 2012, pp. 1--5.

\bibitem{xia2016robust}
Sifeng Xia, Jiaying Liu, Yuming Fang, Wenhan Yang, and Zongming Guo,
\newblock ``Robust and automatic video colorization via multiframe reordering
  refinement,''
\newblock in {\em IEEE International Conference on Image Processing}. IEEE,
  2016, pp. 4017--4021.

\bibitem{yatziv2006fast}
Liron Yatziv and Guillermo Sapiro,
\newblock ``Fast image and video colorization using chrominance blending,''
\newblock {\em IEEE Transactions on Image Processing}, vol. 15, no. 5, pp.
  1120--1129, 2006.

\bibitem{levin2004colorization}
Anat Levin, Dani Lischinski, and Yair Weiss,
\newblock ``Colorization using optimization,''
\newblock in {\em ACM transactions on graphics ({TOG})}. ACM, 2004, vol.~23,
  pp. 689--694.

\bibitem{sheng2014video}
Bin Sheng, Hanqiu Sun, Marcus Magnor, and Ping Li,
\newblock ``Video colorization using parallel optimization in feature space,''
\newblock {\em IEEE Transactions on Circuits and Systems for Video Technology},
  vol. 24, no. 3, pp. 407--417, 2014.

\bibitem{ben2015approximate}
Nir Ben-Zrihem and Lihi Zelnik-Manor,
\newblock ``Approximate nearest neighbor fields in video,''
\newblock in {\em IEEE International Conference on Computer Vision and Pattern
  Recognition (CVPR)}, 2015, pp. 5233--5242.

\bibitem{welsh2002transferring}
Tomihisa Welsh, Michael Ashikhmin, and Klaus Mueller,
\newblock ``Transferring color to greyscale images,''
\newblock in {\em ACM Transactions on Graphics (TOG)}. ACM, 2002, vol.~21, pp.
  277--280.

\bibitem{goodfellow2014generative}
Ian Goodfellow, Jean Pouget-Abadie, Mehdi Mirza, Bing Xu, David Warde-Farley,
  Sherjil Ozair, Aaron Courville, and Yoshua Bengio,
\newblock ``Generative adversarial nets,''
\newblock in {\em Advances in neural information processing systems (NIPS)},
  2014, pp. 2672--2680.

\bibitem{salimans2016improved}
Tim Salimans, Ian Goodfellow, Wojciech Zaremba, Vicki Cheung, Alec Radford, and
  Xi~Chen,
\newblock ``Improved techniques for training {GAN}s,''
\newblock in {\em Advances in neural information processing systems (NIPS)},
  2016, pp. 2234--2242.

\bibitem{daskalakis2017traininggans}
Constantinos Daskalakis, Andrew Ilyas, Vasilis Syrgkanis, and Haoyang Zeng,
\newblock ``Training {GAN}s with optimism,''
\newblock {\em CoRR}, vol. abs/1711.00141, 2017.

\bibitem{isola2016image}
Phillip Isola, Jun-Yan Zhu, Tinghui Zhou, and Alexei~A Efros,
\newblock ``Image-to-image translation with conditional adversarial networks,''
\newblock {\em arXiv preprint arXiv:1611.07004}, 2016.

\bibitem{juliani2017}
A.W. Juliani,
\newblock ``{Pix2Pix-Film},'' \url{http://github.com/awjuliani/Pix2Pix-Film},
  2017,
\newblock [Online; accessed 2-January-2018].

\bibitem{zhang2016colorful}
Richard Zhang, Phillip Isola, and Alexei~A Efros,
\newblock ``Colorful image colorization,''
\newblock in {\em IEEE European Conference in Computer Vision (ECCV)}.
  Springer, 2016, pp. 649--666.

\bibitem{ronneberger2015u}
Olaf Ronneberger, Philipp Fischer, and Thomas Brox,
\newblock ``{U}-{N}et: Convolutional networks for biomedical image
  segmentation,''
\newblock in {\em International Conference on Medical Image Computing and
  Computer-Assisted Intervention}. Springer, 2015, pp. 234--241.

\bibitem{chintala2016}
S.~Chintala, E.~Denton, M.~Arjovsky, and M.~Mathieu,
\newblock ``{How to train a {GAN}? Tips and tricks to make {GAN}s work},''
  \url{http://github.com/soumith/ganhacks}, 2016,
\newblock [Online; accessed 25-January-2018].

\bibitem{dialm}
``Dial {M} for murder,'' https://www.imdb.com/ title/tt0046912/, 1954.

\bibitem{etdieu}
``Et {D}ieu..cr\'{e}a la femme,'' https://www.imdb.com/ title/tt0049189/, 1956.

\bibitem{tzeni}
``Tz\'{e}ni, tz\'{e}ni,'' https://www.imdb.com/ title/tt0145006/, 1966.

\bibitem{astreetcar}
``A streetcar named desire,'' https://www.imdb.com/ title/tt0044081/, 1951.

\bibitem{twelveangrymen}
``Twelve angry men,'' https://www.imdb.com/ title/tt0050083/, 1957.

\end{thebibliography}

\end{document}